\documentclass[]{spie}  

\usepackage{amsmath,amsfonts,amssymb}
\usepackage{graphicx}
\usepackage[colorlinks=true, allcolors=blue]{hyperref}
\usepackage[table,xcdraw]{xcolor}

\title{Quantification of Lung Abnormalities in Cystic Fibrosis using Deep Networks}

\author[1,2]{Filipe Marques}
\author[1]{Florian Dubost}
\author[3]{Mariette Kemner-van de Corput}
\author[3]{Harm A.W. Tiddens}
\author[1,4]{Marleen de Bruijne}
\affil[1]{Biomedical Imaging Group Rotterdam, Department of Radiology and Medical Informatics, Erasmus MC, Rotterdam, The Netherlands}
\affil[2]{Faculty of Engineering of University of Porto, Porto, Portugal}
\affil[3]{Department of Pediatric Pulmonology and Allergology, Erasmus MC Rotterdam, the Netherlands}
\affil[4]{Department of Computer Science, University of Copenhagen, Denmark}

\authorinfo{Further author information: (Send correspondence to F. Marques)\\F. Marques: E-mail: fmmarques16@gmail.com}

\begin{document} 
\maketitle

\begin{abstract}
Cystic fibrosis is a genetic disease which may appear in early
life with structural abnormalities in lung tissues. We propose to detect
these abnormalities using a texture classification approach. Our method
is a cascade of two convolutional neural networks. The first network detects
the presence of abnormal tissues. The second network identifies
the type of the structural abnormalities: bronchiectasis, atelectasis or
mucus plugging.We also propose a network computing pixel-wise heatmaps of abnormality
presence learning only from the patch-wise annotations. Our
database consists of CT scans of 194 subjects. We use 154 subjects to train
our algorithms and the 40 remaining ones as a test set. We compare
our method with random forest and a single neural network approach.
The first network reaches an accuracy of 0,94 for disease detection, 0,18
higher than the random forest classifier and 0,37 higher than the single neural network. Our cascade approach yields a
final class-averaged F1-score of 0,33, outperforming the baseline method and the single network 
by 0,10 and 0,12 .
   
\end{abstract}


\section{INTRODUCTION}
\label{sec:intro}  

\begin{figure}
							\begin{minipage}{0.24\textwidth}
								\raggedleft\includegraphics[width=1.1\textwidth]{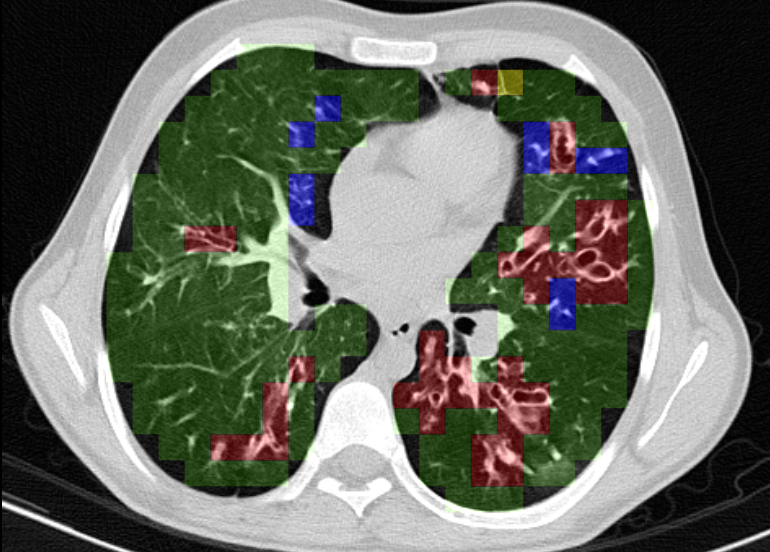}
								
							\end{minipage}
							\hfill
                            \begin{minipage}{0.24\textwidth}
								\raggedleft\includegraphics[width=0.8\textwidth]{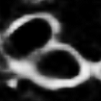}
								
							\end{minipage}
							\hfill
                            \begin{minipage}{0.24\textwidth}
								\centering\includegraphics[width=0.80\textwidth]{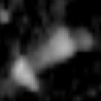}
							
							\end{minipage}
							\begin{minipage}{0.24\textwidth}
								\centering\includegraphics[width=0.8\textwidth]{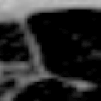}
							
							\end{minipage}

							\caption{Original slice(left) annotated being: Green - Healthy; Blue - Mucus; Red - Bronchiectasis; Yellow - Atelectasis. From left to right example of Bronchiectasis, Mucus and Atelectasis }
							\label{fig:Examples1}
						\end{figure}

 Cystic fibrosis (CF) is the most lethal genetic disorder in the Caucasian population. The disease starts to express itself by changing the structure of lung tissue leading to structural abnormalities \cite{CF}.The early stages of cystic fibrosis have not been thoroughly studied yet. Early, automatic and quantitative analysis could give a better knowledge on what changes lead to Severe Advanced Lung Disease (SALD) \cite{SALD} and reduce irreversible lung damage. The different types of abnormalities studied in this article are bronchiectasis - destruction or widening of the airway - mucus plug
and atelectasis (deflation of alveoli) \cite{PRAGMA} as shown  Fig. \ref{fig:Examples1}.

There are many approaches to automatic lung tissue classification. They can depend on handcrafted or learnt features. Handcrafted features use predefined filter banks containing features that capture uniquenesses and specific details of lung tissue. In Ciompi et al.\cite{Ciompi} for instance, the authors cascade supervised and unsupervised learning based on handcrafted features.

Recently deep learning techniques have successfully been used for medical image analysis. However, few articles \cite{CNN,GAO} propose to use deep learning for texture classification.
A common problem is the unbalanced between different classes that leads to inefficient networks. Wang et al. \cite{Gabor} proposed a CNN where the classes are balanced by oversampling patches of the rare classes.

In this paper we propose an automatic method for patch-wise texture classification of different structural lung abnormalities. Our method is based on a cascade of two convolutional neural networks. To the best of our knowledge it is the first time that structural abnormalities in early stages of cystic fibrosis are automatically scored. Besides, we train and evaluate our algorithm on CT scans of children, acquired with a variety of scanner models and low dose scan protocols. All of this makes the problem more challenging.

In addition we compute precise pixel-wise heatmaps of structural abnormalities \cite{gpunet}, using only the global patch labels during training to visualize where certain abnormalities are most likely to be present.

\section{METHODS}
\label{sec:methods}

We propose a cascade of two convolutional neural networks (CNN) to classify 2D patches from lung CT scans. The first network performs a binary classification between healthy and diseased tissue. The second network refines the prediction by classifying the predicted diseased patches into 4 several specific subgroups: bronchiectasis, atelectasis, mucus or abnormal. As a side result, we were also able to compute pixel-wise heatmaps of abnormality presence from patch-wise annotations.

\subsection{Pre-Processing.}
Only inspiration scans from the subjects are present. From these scans, 2D slices containing annotations were extracted. Annotations are patch-wise with variable size. Patches were resampled to have the same number of voxels. The image intensity of each CT scan was rescaled between 0 and 1. Since the subjects are in a early stage of cystic fibrosis, the changes in lung texture are mild and it can be challenging to distinguish them from healthy tissue. The classified patches are therefore selected larger than the annotations to include anatomical context.


\subsection{Cascade of Convolutional Neural Networks.}
In our dataset, the vast majority of patches are healthy (table \ref{tab:table1}). Direct classification approaches have the tendency to overestimate significantly  the presence of abnormalities to maximize their global learning objective.
To overcome this problem we train two different convolutional neural networks with different loss functions. The first network detects the presence of abnormalities. The second network classifies these abnormalities into several subgroups. 

These two networks have the same architecture except for the last layer. The first network performs a binary classification and ends with a sigmoid activation function. The second network performs a multi-class classification and ends with a softmax function.

\begin{figure}
                            	\begin{minipage}{\textwidth}
								\centering\includegraphics[width=0.9\textwidth]{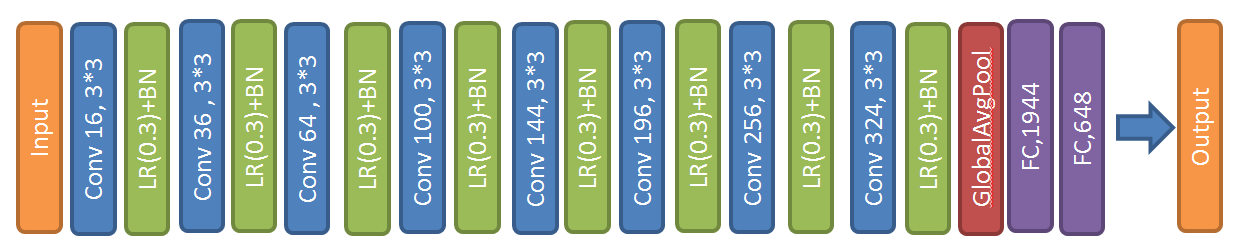}
								\caption{Architecture of both neural networks of our cascade approach. Only the last activation differs: sigmoid/softmax. Blue block - convolutions and the number of kernels associated. Green block - Leaky ReLU and Batch Normalization. Red block - Global Average Pooling. Purple Block - Fully connected layer }
								\label{fig:Arch}
							\end{minipage}
                        \end{figure}

In Fig. \ref{fig:Arch} the network architecture is presented. In this network no pooling is performed between the layers since texture contains low level features that can be lost - there is only global average pooling in the end. The network architecture is similar to the one proposed by Anthimopoulos et al.\cite{CNN} with some adaptations. The size of convolution kernels are 3x3 instead of 2x2. In addition to dropout and data augmentation, a batch normalization layer \cite{BN} is inserted after every convolution . This allowed us to use higher learning rates of the optimizer.

The first network is optimized with a dice coefficient loss function. Being D the dice coefficient, the loss function is L=1-D:
\begin{equation}
L= 1- \frac{2\sum_{i=1}^{n}p_{i}y_{i}}{\sum_{i=1}^{n}p_{i}^{2}+\sum_{i=1}^{n}y_{i}^{2}}
\end{equation}
The second network is optimized with a class-weighted categorical cross-entropy define as: 
\begin{equation}
L=-\frac{1}{n}  \sum_{i=1}^{n}\sum_{j=1}^{m} w_{j} y_{ij} log(p_{ij})
\end{equation}
where $n$ is the number of samples, $m$ the number of classes, $w_{j}$ the weight of class $j$, $y_{ij}$ the ground truth and $p_{ij}$ the prediction. Weighting the cross-entropy compensates, to some extent, the class imbalance in the dataset. As we show in the experiments, this approach alone cannot overcome the significant healthy/disease imbalance of our dataset. 
In order to minimize the loss we used Stochastic Gradient Descent (SGD). To accelerate convergence we use large dense layers.  Before each dense layer a dropout layer prevents over-fitting. For the activation of each layer Leaky ReLu is used in order to avoid the stagnation of a neuron as may happen after a large gradient update \cite{LU}. 

\subsection{Heatmaps of disease presence.} We aim at computing a heatmap of pixel-wise disease presence only using the patch-wise information during
training. This can be seen as a way to automatically refine the precision of the annotations. In this problem
setting, the ground truth patches are considered as weak-labels.

To compute this side result we use a different network from \ref{fig:Arch}. It is a CNN with a U-net architecture \cite{unet} followed by a global pooling layer\cite{nin} based on the network described in Dubost et al. \cite{gpunet}. To compute the heatmap
on the complete axial slice (Fig. 3), we compute the average of the patch-wise heatmaps predictions in the area
where they overlap.

\section{Experiments}
\label{sec:sections}

The evaluation of the method is performed in a test set with 40 CT scans. We compare the performance of the proposed method with a baseline method based on a random forest classifier and a direct application of our CNN on all classes. The two networks from the cascade method are evaluated separately. For detection we evaluate true positive rate (TPR), true negative rate (TNR) and accuracy. For multi-class classification we evaluate the average F1-Score of all classes: harmonic mean between precision and recall. Each network is compared to the baseline method used. In the end of the cascade the results from the two networks are concatenated and evaluated.

\subsection{Data.}
The scans were acquired in Erasmus MC-Sophia and annotated using an in-house developed annotation tool - PRAGMA \cite{PRAGMA}. The dataset consists of 194 scans from children with ages between 1 year and 11 months to 18 years old. The children are in average 9 years old. 144 patients have abnormalities as consequence of CF while the remaining 50 patients don't have annotated abnormalities. The slice thickness ranges between 0,75-3 mm with an average of 1,85 mm.  There is different slice spacing up to 7 mm. The scans were reconstructed with different kernels, B60S and B75f being the most common. All scans were taken at full inspiration breath-hold.
Each scan has 10 annotated slices patch-wise. Each original patch was first resized to 20 by 20 pixels. To incorporate the necessary surrounding information, the area around each patch is included in the inputs to the network. Every patch is extracted to have a final size of 60 by 60 pixels.Number of patches and its distribution is presented in Table \ref{tab:table1}.  

\begin{table}[]
\centering
\caption{Number of patches for each class and the ratio between the presence of the class and the all data}
\label{tab:table1}
\begin{tabular}{c|c|c|c|c|}
\cline{2-5}
\multicolumn{1}{l|}{}         & \multicolumn{4}{c|}{Disease}                     \\ \hline
\multicolumn{1}{|c|}{Healthy} & Bronchiectasis & Abnormal & Mucus  & Atelectasis \\ \hline
\multicolumn{1}{|c|}{217787}  & 2507           & 592      & 622    & 691         \\ \hline
\multicolumn{1}{|c|}{98,00\%} & 1,14\%         & 0,27\%   & 0,28\% & 0,31\%      \\ \hline
                              & \multicolumn{4}{c|}{2\%}                      \\ \cline{2-5} 
\end{tabular}
\end{table}

 The image is overlaid with a square grid. The grid size varies across scans and is defined as one-twentieth of the lung width at the carina (ridge at the base of the trachea). If a grid cell/patch is covered by diseased lung tissue for at least 50\% of its surface, the patch is annotated as either bronchiectasis, mucus plug, atelectasis or abnormal (presence of unusual lung texture not falling into the other categories). Otherwise the patch is annotated as healthy. This leads to a 5 classes patch classification problem.  
In case several diseases are present within the same patch, only one label is assigned. The class is selected according to a hierarchical system, from highest to lowest priority: bronchiectasis, mucus plug,bronchial wall thickening, atelectasis and normal structure.

\subsection{Baseline Methods.}
We compare our method with two competitive non cascaded approaches: a single/direct CNN with the same architecture as our networks and a random forest classifier with features similar to the one presented in Ciompi et al. \cite{Ciompi}. For the random forest classifier, given a 60x60 voxel-wide patch, its feature vector $x$ is computed as follows. We first filter the patch with 15 different features based on Gaussian filter, Gradient Magnitude, Laplacian and eigenvalues of Hessian matrix, all with different kernel sizes ($\sigma_1=0.5$, $\sigma_2=1$,$\sigma_3=1.5$). We then compute 16 different intensity histograms from these 15 filtered patches plus the original intensities. Each histogram has 100 bins and histogram equalization is subsequently computed. These 16 histograms are then concatenated into a single feature vector $x \in \mathbb{R}^{1600}$.

\subsection{Experimental Settings.}
The algorithms are implemented in Python with Keras and Theano libraries. The experiments ran on a Nvidia GeForce GTX 1070 GPU. Data processing was done in MatLab and Python. The class-weights in the cross-entropy loss function of the second network are the following: bronchiectasis 1.2, abnormal 1, mucus 1.8 and atelectasis 1.8. For the direct CNN baseline method the disease weights are the same and the weight for healthy class is 0.005. Mucus and atelectasis occur less frequently, hence have higher weight. The abnormal class has
a low intra-observer agreement and a low clinical priority, therefore we associate a lower weight to this class.
To compensate the healthy/disease class imbalance, the disease patches are replicated 16 times for the training
of the first network. This leads to a 1/4 disease/healthy balance. As our batch size is 64, both labels are, on
average, present in a batch.

\subsection{Results.} Table \ref{tab:table2} compares the class-average F1-score (as defined in \cite{CNN}) between the random forest, the direct CNN and the cascaded CNNs. The cascaded CNNs outperform the random forest by 0,10  and the direct CNN by 0,12.

\begin{table}[h!]
  \centering
  \caption{Comparison of F-Average Score }
  \label{tab:table2}
  \begin{tabular}{c|c|c}
    Random Forest & Direct CNN & Cascaded CNNs\\
    \hline
    0,23 & 0,21 & 0,33\\
  \end{tabular}
\end{table}

\begin{figure}
\centering
\includegraphics[width=0.9\textwidth]{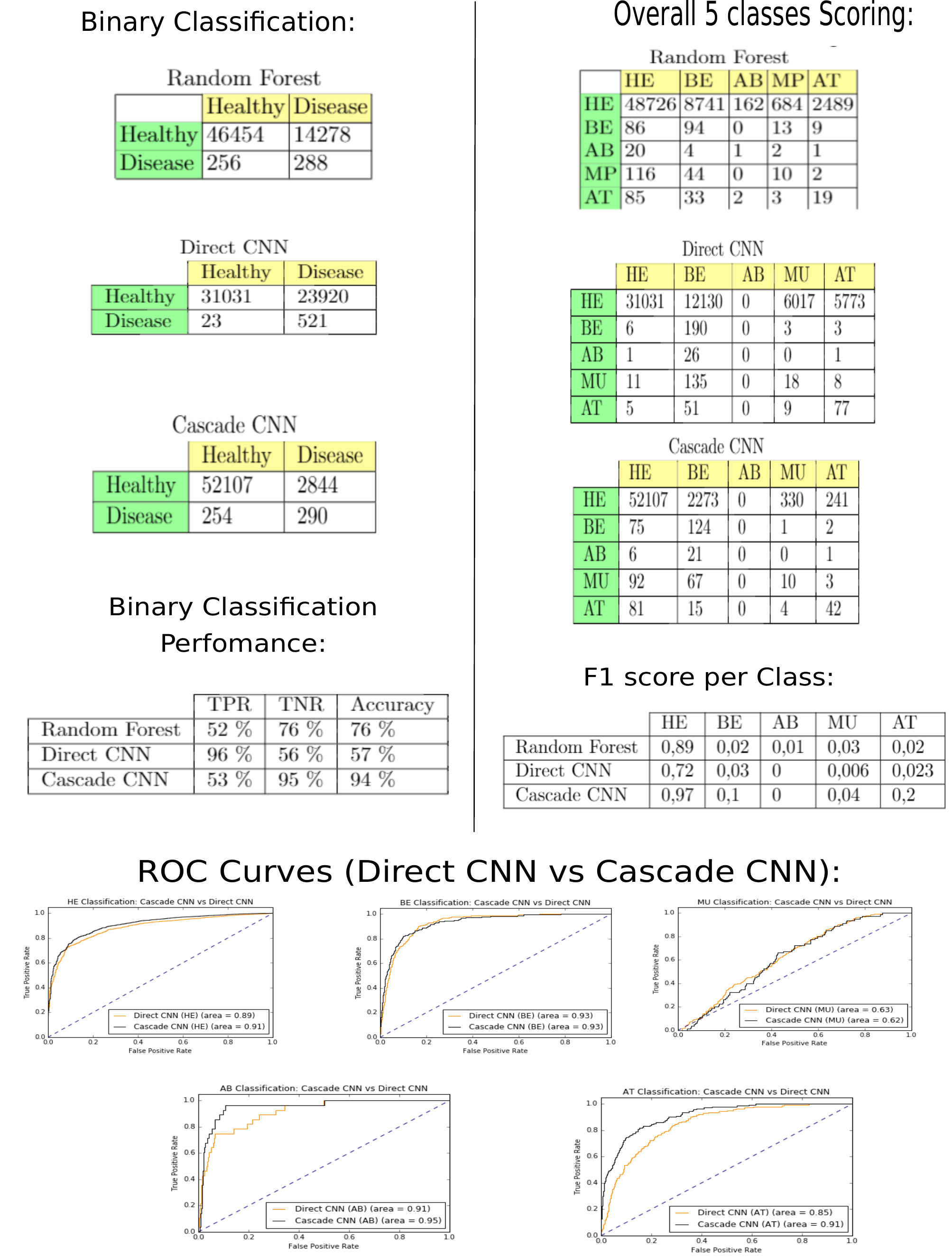}
\caption{Results for each approach. Yellow: Predicted Class; Green:Ground-Truth; Left: binary classification for disease detection; Right: Scoring of all classes present. Classes: HE - Healthy; BE-Bronchiectasis; AB - Abnormal; MP- Mucus Plugging; AT - Atelectasis. Metrics: TPR - True Positive Rate; TNR - True Negative Rate.  For the random forest, the patients with no disease patches were separated differently, which resulted in a different number of healthy patches than for the direct and cascaded CNN.
Bottom Part: ROC Curves for each disease and healthy for Direct network and Cascade Network.}
\label{fig:confusion}
\end{figure}

In Fig. \ref{fig:confusion} we detailed the classification for every class and compare with the baseline methods.
For the binary healthy/disease classification, in order to compare the direct CNN with cascaded CNNs, all disease predictions of the direct CNN are grouped together in the disease class. Cascaded CNNs greatly outperform the two other approaches in every metric. 

For abnormality scoring, direct CNN and cascade CNN show similar performances. Cascade CNN fails at scoring the abnormal class. This happens because abnormal class represents an unusual structure that does not resemble  the other diseases. As mentioned before abnormal class is poorly annotated by the clinicians. In the intra-observer agreement this class presents an intraclass correlation of 0,13. Therefore this class is not included in the class-averaged of F1-score reported in table \ref{tab:table2}.
In the confusion matrices for disease classification, mucus appears - for all methods - to often be predicted as bronchiectasis. This can be explained by the fact that mucus plugging often coexists with bronchiectasis, filling dilated airways, in which case the PRAGMA score would label this as bronchiectasis.  

Cascade CNN outperforms all other approaches for the overall classification. The gap in average F1-score between cascade CNN and other approaches is due mainly to an accurate detection of disease in the binary classification, reducing significantly the number of false detections in comparison with the other methods.

\begin{figure}

								\includegraphics[width=\textwidth]{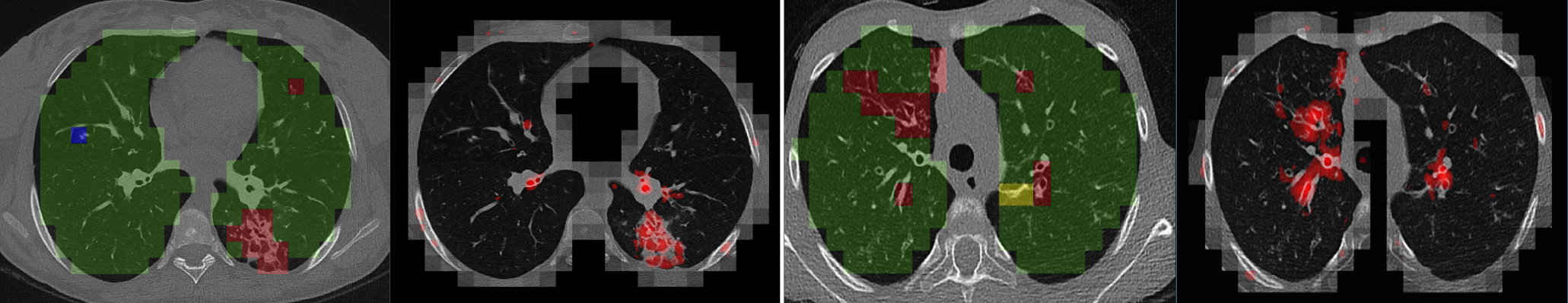}

							\caption{Original slices(left) with annotations and heat-map of abnormality on the same slice (right) Green - Healthy; Red - Bronchiectasis; Blue - Mucus}
							\label{fig:Examples}
						\end{figure}

In Fig. \ref{fig:Examples}, we show some examples of the pixel-wise heatmaps of disease presence computed as explained in section \ref{sec:methods}. The heatmaps are thresholded to highlight the strongest predictions. Bronchietasis is localized quite accurately. It seems more difficult to detect atelectasis and mucus. This might be because of the imbalance of the data. As mentioned in the data section, bronchiectasis is annotated in our dataset with the highest (clinical) priority. This may have introduced a bias leading to an overestimation of this class. In Fig. \ref{fig:confusion}, all approaches overestimate bronchiectasis, leading to a high number of false positive and lowering the F1-score for bronchiectasis.

\section{Discussion and Conclusion}
Our results show that the proposed method is able to detect abnormality and
estimate a score for each disease in early stage of CF lung disease. The computed
heatmaps highlight abnormal regions and provide more precise quantification.
The random forest approach presented in Ciompi et al. \cite{Ciompi} showed a good performance in
case of severe advanced lung disease \cite{Ciompi} but is outperformed by our method in data with only mild disease.

Our network is similar to the one in Anthimopoulos et al. \cite{CNN} Both networks have a similar architecture and
are trained for multi-class texture classification in lungs. In Anthimopoulos et al. \cite{CNN}
the network is designed for
detection of irregularities in the pulmonary interstitium, while our network is trained for classification of early signs of structural lung damage in cystic fibrosis. These two problems are different and present different challenges.

We used a 2D approach. The use of 3D convolutions could be better in the thin-slice scans, however because
of the large variation in slice spacing up to 7 mm in our data, we opted for 2D convolutions.

We proposed a cascade method of two convolutional neural networks for lung texture classification in early
stages of cystic fibrosis. The method combines a binary classification to discriminate between healthy and abnormal lung tissue and a second network
that performs a multi-class classification to score different different types of abnormalities. Our method outperforms the baseline method by 0,10 of F-score. We also propose to compute pixel-wise abnormality
maps, only using patch-wise information for training. This can be considered as a way to refine the manual annotations
and circumvent the ambiguity inherent to patch-wise annotations.

\end{document}